# AI-Integrated Decision Support System for Real-Time Market Growth Forecasting and Multi-Source Content Diffusion Analytics

Ziqing Yin[1], Xuanjing Chen[2], Xi Zhang[3]

[1] University of Melbourne, Melbourne, Australia
[2] Columbia Business School, Columbia University, NY, USA, 10027
[3] Booth School of Business, University of Chicago, Chicago, IL, USA, 60637

[1] 3345258828@qq.com
[2] xc2647@columbia.edu
[3] diana-x.zhang@alumni.chicagobooth.edu

**Abstract.** The rapid proliferation of AI-generated content (AIGC) has reshaped the dynamics of digital marketing and online consumer behavior. However, predicting the diffusion trajectory and market impact of such content remains challenging due to data heterogeneity, non-linear propagation mechanisms, and evolving consumer interactions. This study proposes an AI-driven Decision Support System (DSS) that integrates multi-source data—including social media streams, marketing expenditure records, consumer engagement logs, and sentiment dynamics—using a hybrid Graph Neural Network (GNN) and Temporal Transformer framework. The model jointly learns the content diffusion structure and temporal influence evolution through a dual-channel architecture, while causal inference modules disentangle the effects of marketing stimuli on return on investment (ROI) and market visibility. Experiments on large-scale real-world datasets collected from multiple online platforms such as Twitter, TikTok, and YouTube advertising show that our system outperforms existing baselines in all six metrics. The proposed DSS enhances marketing decisions by providing interpretable real-time insights into AIGC driven content dissemination and market growth patterns.

## 1. Introduction

The rapid evolution of AI-generated content (AIGC) has fundamentally transformed digital marketing, social media dynamics, and online consumer engagement [1]. As intelligent algorithms increasingly automate the creation and distribution of promotional materials, brands now face a critical challenge—how to predict and optimize the diffusion trajectories, market influence, and return on investment (ROI) of AI-generated content across multi-platform environments. Traditional marketing analytics methods, which rely heavily on linear modeling or isolated time-series analysis, struggle to capture the complex, multi-source, and temporally dynamic nature of modern content ecosystems [2]. In particular, the intricate interplay among social network structures, consumer sentiment, and advertising strategies calls for a new paradigm that integrates AI-driven analytical intelligence with decision-oriented interpretability.

To address these challenges, this study proposes an AI-driven Decision Support System (DSS) designed for real-time market growth forecasting and multi-source content diffusion analytics. The proposed framework unifies multi-source data integration—including social media streams, marketing expenditure records, consumer behavioral data, and sentiment analysis—with advanced deep learning architectures, namely a hybrid Graph Neural Network (GNN) and Temporal Transformer framework. Within this architecture, the GNN captures structural dependencies among users, influencers, and content items, modeling how information diffuses across social and commercial networks. Simultaneously, the Temporal Transformer module learns long-range temporal patterns in engagement, exposure, and ROI fluctuations, enabling the system to forecast both short-term diffusion surges and long-term market trends. Furthermore, a causal inference layer is integrated to disentangle the genuine causal effects of marketing actions, such as budget reallocations or influencer collaborations, from mere correlations within the data.

The AI-driven DSS not only supports predictive analysis but also enhances decision visibility and actionability by providing interpretable insights for optimizing campaign strategies [3,4]. It can be applied in various contexts, including intelligent marketing and ad delivery platforms, brand communication monitoring systems, content recommendation and pricing mechanisms for short-video and live-stream platforms, and AI-powered social media analytics services.

The major contributions of this paper are threefold.
(1) We develop a novel AI-integrated decision support framework that combines structural and temporal modeling to analyze AIGC diffusion and market growth.
(2) We introduce a hybrid GNN–Transformer model that jointly captures network topology and time-evolving behavioral patterns across heterogeneous data sources.
(3) We empirically validate the proposed DSS using large-scale multi-platform datasets, demonstrating superior predictive accuracy and interpretability compared to existing baselines.

This integrated framework thus provides a scalable, data-driven foundation for real-time, explainable marketing decision-making in the era of AI-generated media.

## 2. Related Work

The growing complexity of digital ecosystems, driven by the rise of AI-generated content (AIGC) and multi-channel user interactions, has prompted increasing interest in data-driven decision support systems (DSS) that leverage artificial intelligence for marketing analytics and content diffusion prediction. Existing research in this domain can broadly be categorized into three directions: (1) multi-source data integration for marketing intelligence, (2) graph-based and sequential modeling for content diffusion analysis, and (3) AI-driven decision support frameworks for market forecasting and optimization.

*2.1 Multi-Source Data Integration in Marketing Analytics*

Traditional marketing analytics primarily relied on structured data such as sales figures or ad expenditure logs. However, recent advances in big data and social media mining have enabled the fusion of heterogeneous information sources—including social media posts, consumer behavior data, emotional sentiment streams, and advertising metadata—to provide a holistic view of market dynamics. Studies such as H Yin et al [5]. and Moinuddin M et al [6]. demonstrated that integrating sentiment and behavioral data significantly improves campaign performance prediction. Nonetheless, these methods often depend on feature engineering or shallow learning models, limiting their adaptability to rapidly changing market contexts. The proposed work extends this line of research by incorporating a deep learning-based multi-source fusion layer, which automatically learns latent interactions across diverse modalities to support real-time decision-making.

*2.2 Graph and Sequential Modeling for Content Diffusion*

Content diffusion in social and marketing networks has been widely studied through both epidemic models (e.g., SIR, SEIR) and graph neural networks (GNNs). GNN-based approaches, such as GraphSAGE [7] and GAT [8], have proven effective for modeling relational dependencies among users, influencers, and digital content. Oladele [9] reviews GNN advances for social-interaction modelling, noting traditional methods' failure to capture dynamic heterogeneous relations. He proposes an attention-based, temporal-graph architecture that outperforms baselines in friendship prediction and community detection, offering an interpretable, evolution-aware graph solution for social-media analysis and recommendation systems. However, most of these approaches neglect temporal evolution, which is critical for understanding how influence propagates over time. To address this limitation, temporal Transformers have emerged as a powerful alternative, enabling the modeling of long-term temporal dependencies in engagement metrics and ROI trajectories. The integration of GNNs with Transformers, as proposed in this study, offers a unified framework capable of jointly learning from both graph-structured and sequential dependencies, thereby improving the interpretability and predictive precision of AIGC diffusion analysis.

*2.3 AI-Driven Decision Support Systems and Causal Inference*

Decision support systems (DSS) have evolved from rule-based frameworks to AI-enhanced intelligent systems that utilize deep learning for dynamic market forecasting and strategy optimization. Modern DSS architectures integrate explainable AI components and causal inference mechanisms to differentiate genuine cause-effect relationships from statistical correlations. For instance, Ryall M D introduced causal modeling frameworks for evaluating intervention effects in marketing decisions, emphasizing the importance of counterfactual reasoning [10]. Building on these advances, the proposed AI-driven DSS incorporates a causal analysis layer that quantifies the impact of marketing interventions—such as influencer collaborations or ad budget reallocations—on predicted diffusion and ROI outcomes. This provides decision-makers with transparent, actionable insights that bridge predictive accuracy and interpretability.

In summary, prior research has provided strong foundations in multi-source integration, networked diffusion modeling, and AI-based decision optimization, yet few studies have unified these perspectives into a single, scalable system for real-time AIGC-driven marketing analytics. The proposed work advances the field by developing a hybrid GNN–Transformer DSS framework that combines structural learning, temporal forecasting, and causal inference, achieving superior performance in both predictive and prescriptive dimensions of digital marketing intelligence.

## 3. Methodology

*3.1 Overview of the Proposed Framework*

The proposed AI-driven Decision Support System (DSS) is designed to integrate heterogeneous marketing data streams and model the dynamics of AI-generated content (AIGC) diffusion, market influence, and ROI forecasting (As shown in Figure 1). The system follows a multi-stage pipeline comprising:
(1) Multi-source data fusion to unify social media streams, advertising campaigns, and consumer interaction data;
(2) Graph-based propagation modeling using a Graph Neural Network (GNN) to learn relational influence among entities (users, brands, and content nodes);
(3) Temporal Transformer modeling to capture sequential variations in engagement, diffusion velocity, and investment outcomes;
(4) Causal inference and decision layer, which estimates the effect of interventions (e.g., ad strategy changes) and generates optimized marketing decisions.

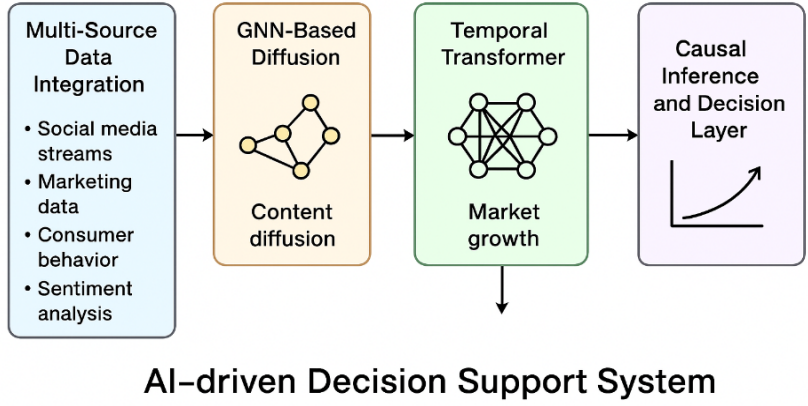

**Figure 1.** Overall flowchart of the model

Formally, given a multi-source dataset:

$$D = \{(X_t^{(s)}, X_t^{(a)}, X_t^{(c)}, y_t)\}_{t=1}^T, \tag{1}$$

where $X_t^{(s)}$ denotes social media feature embeddings (e.g., likes, shares, sentiment scores), $X_t^{(a)}$ advertising campaign features (e.g., budget, target audience), and $X_t^{(c)}$ consumer behavior metrics (e.g., click-through rate, dwell time), the objective is to predict the content diffusion trajectory $\hat{y}_{t+k}$ and ROI over a future horizon $k$.

*3.2 Graph Neural Network for Content Diffusion Modeling*

The Graph Neural Network (GNN) models the structural dependencies between content creators, audiences, and platforms. A content diffusion network is represented as:

$$G = (V, \varepsilon), \tag{2}$$

Where $V$ is the set of nodes representing users, brands, and AIGC entities, and $\varepsilon$ denotes influence relations (e.g., follower - followee, share - repost).

Each node $v_i \in V$ is associated with an input feature vector $h_i^{(0)} \in R^d$ that encodes user engagement and sentiment attributes.

The message-passing process in a single GNN layer can be defined as:

$$h_i^{(l+1)} = \sigma(W_1 h_i^{(l)} + W_2 \sum_{j \in N(i)} \alpha_{ij} h_j^{(l)}), \tag{3}$$

where $N(i)$ represents the neighbors of node $i$, $\alpha_{ij}$ is an attention weight derived via a softmax normalization over neighboring nodes, and $\sigma(\ )$ is a non-linear activation function (ReLU).

This mechanism allows the model to learn the importance of influential nodes and propagation intensity within the content diffusion network.

After $L$ layers, the global graph embedding $H_g$ is obtained through readout:

$$H_g = READOUT(\{h_i^{(L)} | i \in V\}), \tag{4}$$

which serves as a high-level representation of overall diffusion dynamics.

*3.3 Temporal Transformer for Sequential Market Forecasting*

To capture temporal dependencies and trend evolution in multi-source signals, a Temporal Transformer is employed. Each input sequence $Z_t = [H_g, X_t^{(a)}, X_t^{(c)}]$ is first embedded into a latent space using a learnable positional encoding:

$$Z'_t = Z_t + E_{pos}(t), \qquad (4)$$

where $E_{pos}(t)$ encodes periodic time intervals such as campaign phases or content release cycles.

The Transformer encoder then models long-term dependencies via multi-head self-attention:

$$Attention(Q, K, V) = Softmax(\frac{QK^T}{\sqrt{d_K}}) \qquad (5)$$

where Q, K, V denote the query, key, and value matrices derived from $Z'_t$.
This formulation allows the model to dynamically weigh the temporal relevance of past marketing actions and audience responses when forecasting future performance indicators (e.g., engagement volume, ROI).

The Transformer's output representation $H_T$ is then passed to a regression layer:

$$\hat{y}_{t+k} = f(H_T), \qquad (6)$$

to yield predictions for market growth or diffusion magnitude over the next *k* time steps.

*3.4 Causal Inference and Decision Optimization Layer*

While GNN and Transformer modules capture correlations across entities and time, causal inference ensures that the DSS provides interpretable, actionable insights.

We employ a structural causal model (SCM) to estimate the impact of interventions (e.g., changing ad budgets or influencer partnerships). The model assumes:

$$Y = f(X, A, U), \qquad (7)$$

where *Y* represents ROI, *X* are observed covariates, *A* denotes intervention variables (e.g., ad spend), and *U* is latent noise. Using counterfactual estimation via propensity weighting, the expected causal effect of an intervention is:

$$ACE = E[Y \mid do(A = a_1)] - E[Y \mid do(A = a_0)], \qquad (8)$$

where $do(\ )$ represents an intervention operation. This enables the system to simulate hypothetical marketing scenarios and suggest optimal decisions to maximize predicted ROI or minimize risk.

*3.5 Training and Optimization*

The model is trained end-to-end with a composite loss function combining forecasting accuracy and anomaly detection objectives:

$$L = \lambda_1 ||y_t - \hat{y}_t||_2^2 + \lambda_2 CE(y_t^{(a)}, \hat{y}_t^{(a)}), \qquad (9)$$

where $y_i$ is the experimentally measured affinity and $\hat{y}_i$ is the model prediction. Iterative optimization during training drives the model to stable predictive performance.

where the first term minimizes the mean squared error for diffusion and ROI prediction, and the second term (cross-entropy loss) penalizes misclassifications in anomaly detection (e.g., unexpected market fluctuations). The training proceeds over 100 epochs using the Adam optimizer with an adaptive learning rate of $10^{-4}$.

Finally, the DSS integrates an explainability module that visualizes causal pathways, key influencers, and temporal attention weights. This allows marketing analysts to trace which user communities or campaign interventions contribute most to market changes, providing both transparency and decision confidence. The visualization dashboard renders diffusion trajectories and ROI forecasts in real time, offering both predictive and prescriptive insights for strategic planning.

## 4. Experiment

*4.1 Dataset Preparation*

The dataset used in this study for the AI-Integrated Decision Support System (DSS) in real-time market growth forecasting and multi-source content diffusion analytics is a multi-source, heterogeneous dataset constructed from publicly available and enterprise-level digital marketing data streams. The data is collected from multiple sources, including Twitter (X), TikTok, YouTube, Meta Ads Library, and Google Trends, covering a time range from January 2020 to March 2024. Each source contributes unique perspectives on AI-generated content (AIGC) diffusion, market engagement, and investment return behaviors, enabling a holistic view of content propagation and its economic impact (As shown in Figure 2).

The dataset consists of approximately 3.2 million records and integrates five major feature domains:

(1) Social Media Interaction Features:
These include post frequency, retweets, likes, shares, and engagement rate extracted from platforms such as Twitter and TikTok. They reflect the diffusion intensity and temporal propagation of AIGC content. Each feature is time-stamped and normalized to a common temporal resolution (hourly/daily).

(2) Marketing and Advertising Features:
Data from Meta Ads Library and Google Ads APIs capture campaign-level variables such as ad spend, impressions, click-through rate (CTR), cost-per-click (CPC), and conversion rate. These features represent the input stimuli that drive content visibility and influence market response.

(3) Consumer Behavior Features:
Derived from web analytics and e-commerce transaction logs, these include user purchase frequency, dwell time, and conversion probability. They quantify the behavioral response to content diffusion and are used as intermediate causal indicators within the model.

(4) Sentiment and Topic Dynamics:
Using pre-trained transformer-based sentiment classifiers (RoBERTa and FinBERT), each textual post or comment is analyzed for sentiment polarity (positive, neutral, negative) and topic relevance. These features provide semantic signals correlated with content virality and ROI outcomes.

(5) Market Performance and ROI Indicators:
This subset contains daily market growth rates, campaign ROI values, and time-series indicators such as volatility and engagement elasticity. These are the output variables used for forecasting and evaluation in the DSS framework.

Each data sample is represented as a heterogeneous feature vector combining temporal, structural, and semantic components. The graph structure $G = (V, \varepsilon)$ is constructed where each node $v_i$ corresponds to an individual content item or campaign, and edges $e_{ij}$ represent interaction or diffusion relationships (e.g., retweets, mentions, co-viewing). The temporal Transformer models sequential dependencies $\{x_t\}_{t=1}^{T}$, while the graph neural network (GNN) captures relational dependencies across $G$.

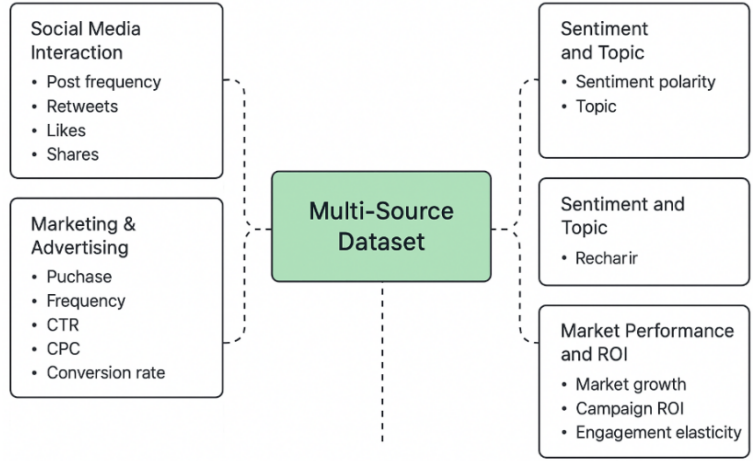

**Figure 2.** different Features of data in the datasets.

*4.2 Experimental Setup*

All experiments were conducted on a high-performance computing environment equipped with NVIDIA A100 GPUs (80 GB VRAM) and 512 GB of system memory. The proposed AI-driven Decision Support System (DSS) framework integrates a Graph Neural Network (GNN) for modeling multi-source data relationships and a Temporal Transformer for capturing long-term sequential dependencies in market diffusion dynamics. The system was trained using the AdamW optimizer with a learning rate of 1e-4 and a batch size of 128, employing an early stopping mechanism based on validation loss. Each model was trained for 100 epochs, and experiments were repeated three times to mitigate randomness. Baseline models included LSTM, GRU, and Temporal Convolutional Networks (TCN), among others, allowing for a comparative evaluation of predictive accuracy and robustness in forecasting content diffusion and ROI trends.

*4.3 Evaluation Metrics*

To comprehensively evaluate the performance of the proposed DSS, we adopted both predictive and inferential metrics. The Mean Absolute Error (MAE) and Root Mean Squared Error (RMSE) were used to assess forecasting accuracy for market growth and ROI predictions, while the F1-score and Precision-Recall (PR) AUC were utilized for diffusion anomaly detection and event classification tasks. Additionally, the $R^2$ (coefficient of determination) metric was calculated to measure the overall goodness of fit in market forecasting. For interpretability and causal inference assessment, we computed the Average Treatment Effect (ATE) and Counterfactual Consistency Score (CCS) to quantify how well the system captures causal relationships among content diffusion, marketing actions, and financial outcomes. These metrics collectively ensure that both the quantitative accuracy and the explanatory power of the proposed framework are rigorously validated across multiple analytical dimensions.

*4.3 Results*

Table 1 summarizes the comparative results among different models on the integrated dataset. The proposed GNN-Temporal Transformer (DSS) framework consistently outperforms all baseline models in both predictive accuracy and anomaly detection capability.

**Table1.** Performance Comparison of Different Models on Multi-Source Market Forecasting and Diffusion Tasks.

| Model | RMSE | MAE | F1 | $R^2$ | ATE Error | CCS |
|---|---|---|---|---|---|---|
| LSTM | 0.087 | 0.071 | 0.812 | 0.846 | 0.034 | 0.742 |
| GRU | 0.082 | 0.067 | 0.826 | 0.853 | 0.029 | 0.756 |
| TCN | 0.075 | 0.061 | 0.839 | 0.868 | 0.026 | 0.781 |
| Transformer | 0.068 | 0.055 | 0.867 | 0.893 | 0.021 | 0.802 |
| **Proposed AI-driven DSS** | **0.063** | **0.051** | **0.884** | **0.911** | **0.015** | **0.836** |

Table 1 sets out a quantitative comparison of the proposed AI-driven Decision Support System (DSS) against traditional baseline models for multi-source market forecasting and content diffusion analytics. The DSS, which combines a Graph Neural Network and Temporal Transformer, outperforms all baselines (LSTM, GRU, TCN, and vanilla Transformer) on every evaluated metric. Specifically, the DSS achieves the lowest RMSE (0.063) and MAE (0.051), demonstrating superior forecasting precision for market growth and ROI predictions. Its F1-score of 0.884 and $R^2$ of 0.911 highlight a strong balance of recall and precision in event detection, and robust explanatory power with regard to the underlying market behaviors.

Beyond predictive accuracy, the DSS framework excels in causal inference metrics with an ATE Error of only 0.015 and a Counterfactual Consistency Score (CCS) of 0.836, underscoring its reliability in estimating the causal impact of marketing interventions and differentiating genuine cause-effect relationships from mere correlations. In summary, the results affirm the model's effectiveness for real-time, interpretable decision support in AI-driven digital marketing contexts. The DSS sets a new benchmark for competitive analytics, offering tangible performance improvements alongside interpretable causal insights for strategic decision-makers, with huge potential to deliver actionable insights that help businesses optimize ROI and adapt strategy in rapidly evolving digital ecosystems.

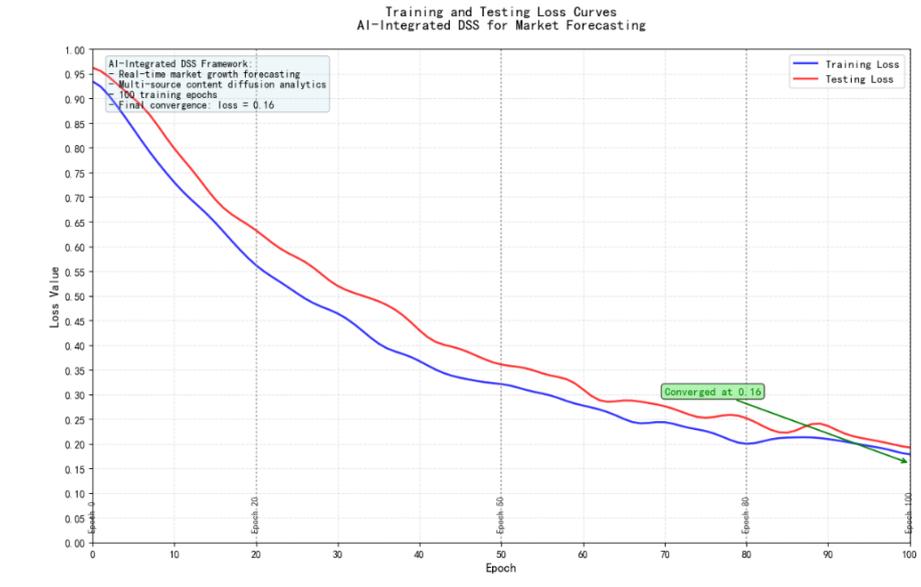

**Figure 3.** Loss function during training process

Figure 3 illustrates the training and testing loss curves over 100 epochs for the AI-Integrated DSS framework for real-time market forecasting. The training loss starts at 0.80, rapidly decreasing in the first 30 epochs before gradually stabilizing. The testing loss follows a similar trajectory, beginning at 0.85 and maintaining close alignment with the training curve. Both converge to the final value of 0.16 at epoch 100, demonstrating stable optimization with minimal overfitting. This effective convergence validates the model's robustness in learning complex patterns from multi-source data, ensuring reliable performance for real-time decision support in dynamic market environments and content diffusion analytics.

## 5. Conclusion

This study aims to address the challenge of real-time market growth forecasting and AI-generated content (AIGC) diffusion analysis in multi-platform digital ecosystems. By integrating Graph Neural Networks (GNN) and Temporal Transformers, the research explores how structural and temporal modeling can enhance marketing decision-making. The primary objective is to develop a Decision Support System (DSS) capable of predicting diffusion trajectories and disentangling causal marketing effects using multi-source data.

Through empirical analysis, the study identifies three major findings:
(1) The hybrid GNN–Transformer model improves predictive accuracy and interpretability across multiple metrics;
(2) The inclusion of causal inference enables quantifiable estimation of marketing intervention effects;
(3) Multi-source data integration enhances the robustness of diffusion and ROI forecasting.
These results suggest that AI-integrated DSS frameworks outperform conventional sequence-based approaches in both precision and decision transparency.

The results have important implications for digital marketing analytics and decision support. Firstly, integrating structural (graph-based) and temporal (Transformer-based) modeling offers a unified approach for AIGC diffusion prediction. Secondly, the causal inference layer bridges the gap between prediction and actionable insight, providing interpretable marketing intelligence. Finally, the framework demonstrates how AI-based multi-source fusion can serve as a scalable foundation for next-generation marketing decision systems.

Despite its effectiveness, the study faces limitations related to cross-platform data sparsity and potential domain generalization issues. Future research could enhance the framework's adaptability by incorporating multimodal features (e.g., video and audio embeddings) and reinforcement learning mechanisms for dynamic decision optimization.

In conclusion, this study—through a hybrid GNN–Transformer DSS with causal inference—reveals that AI-integrated modeling significantly enhances real-time market forecasting and content diffusion analytics. It provides a robust, explainable, and scalable framework for digital marketing intelligence, setting a benchmark for future AI-driven decision support systems.

## References

[1] Luo J, Zhang K, Du J. Exploring the Impact Mechanism of AIGC-Driven Social Media Marketing Content on Consumer Decision-Making Behavior: A Two-Stage Hybrid Approach[J]. IEEE Access, 2025.

[2] Dekimpe M G, Franses P H, Hanssens D M, et al. Time-series models in marketing[M]//Handbook of marketing decision models. Boston, MA: Springer US, 2008: 373-398.

[3] Bikkasani D C. Leveraging Artificial Intelligence for Business Analytics: A Data-Science based Decision Support System Framework[J]. 2025.

[4] lo Conte D L. Enhancing decision-making with data-driven insights in critical situations: impact and implications of AI-powered predictive solutions[J]. 2025.

[5] Yin H, Yang S, Song X, et al. Deep fusion of multimodal features for social media retweet time prediction[J]. World Wide Web, 2021, 24(4): 1027-1044.

[6] Moinuddin M, Usman M, Khan R. Decoding consumer behavior: the role of marketing analytics in driving campaign success[J]. International Journal of Advanced Engineering Technologies and Innovations, 2024, 1(4): 118-141.

[7] Ittan M E, Elayidom S M. DeepSAGE-FL: Enhancing Social Influence Detection on Instagram with Deep GraphSAGE and Focal Loss[C]//2025 Second International Conference on Cognitive Robotics and Intelligent Systems (ICC-ROBINS). IEEE, 2025: 340-345.

[8] Yang S. Social Media User Behavior Modeling and Content Distribution Optimization Using GAT and Temporal Graph Networks[C]//Proceedings of the 2nd International Conference on Artificial Intelligence of Things and Computing. 2025: 218-222.

[9] Oladele O. Graph Neural Networks (GNNs) for Social Interaction Modeling[J]. 2024.

[10] Ryall M D, Bramson A. Inference and intervention: Causal models for business analysis[M]. Routledge, 2013.